\documentclass[letterpaper]{article}
\usepackage{iccc}

\usepackage{graphicx}
\usepackage{amsfonts}
\usepackage{xcolor}
\usepackage{multirow}

\usepackage{fvextra}
\usepackage[most]{tcolorbox}
\usepackage{enumitem}
\usepackage{times}
\usepackage{helvet}
\usepackage{courier}
\title{World-State Transformations for Neuro-symbolic Interactive Storytelling}

\author{Santiago Góngora$^1$ \qquad {\bf Luis Chiruzzo}$^1$\qquad {\bf Gonzalo Méndez}$^2$\qquad {\bf Pablo Gervás}$^2
$\\\\
$^1$ Instituto de Computación, Facultad de Ingeniería, Universidad de la República, Uruguay \\ 
$^2$ Facultad de Informática, Universidad Complutense de Madrid, Spain
}

\begin{document} 
\maketitle
\begin{abstract}
\begin{quote}
Large Language Models (LLMs) have changed the possibilities of Interactive Storytelling systems that process free-text user input. 
However, as more of these systems are built, evidence continues to mount regarding the story coherence problems that arise when relying solely on them. 
Recent research suggests that LLMs can effectively predict state changes within rule-based Interactive Storytelling systems, triggering pre-programmed \textbf{world-state transformations}. 
In this paper, we conduct an exploratory evaluation of whether such \textit{transformations} can serve as a catalyst for player expression while aiming to address the incoherence issues typical of purely LLM-based approaches. 
Building upon a neuro-symbolic architecture, we conducted experiments using an open-source model (Llama 3 70B) and a closed-source model (Gemini 1.5 Flash), with testing conducted in both English and Spanish. 
Eight participants played two scenarios, carefully designed to assess different evaluation objectives.
Our observations suggest that \textit{transformations} offer a way to maintain world-state consistency while encouraging players to interact creatively through their written inputs.

\end{quote}
\end{abstract}

\section{Introduction}

\newcommand{\citesubject}[1]{\citeauthor{#1}~(\citeyear{#1})}
\newcommand{\todo}[1]{\noindent \textcolor{violet}{\textbf{\small [TODO: #1]}}\\}
\DefineVerbatimEnvironment{PromptVerbatim}{Verbatim}{
  fontsize=\scriptsize,
  breaklines=true,
  breakanywhere=true,
  commandchars=\\\{\}
}
\newtcolorbox{promptbox}[1]{
  enhanced,
  breakable,
  colback=white,
  colframe=black,
  boxrule=0.5pt,
  arc=2pt,
  left=6pt,right=6pt,top=6pt,bottom=6pt,
  title=\textbf{#1},
  fonttitle=\small,
}

Interactive Storytelling (IS) is a broad term for any system that allows users to influence how a story develops~\cite{survey_digital_storytelling}.
These systems face a constant struggle to balance player agency (i.e., the level of control over the story the user has) with a coherent plot, as giving the user more control often compromises the story’s consistency~\cite{Riedl_Bulitko_2012}

\begin{figure}
    \centering
    \includegraphics[width=\linewidth]{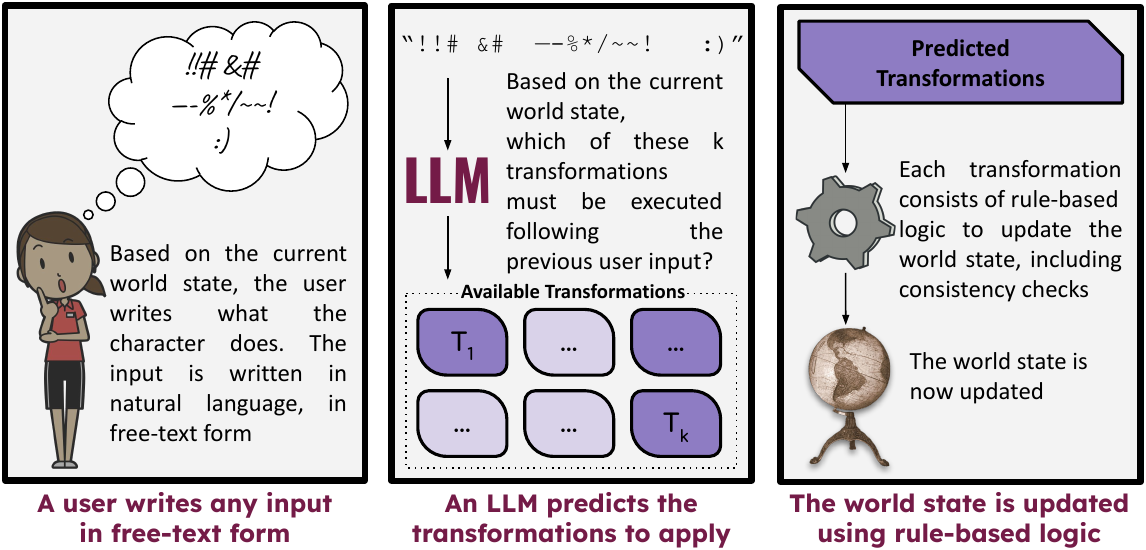}
    \caption{An overview of how \textit{transformations} can be used to tackle the world-update problem.}
    \label{fig:transformations_overview}
\end{figure}

One of the most prominent types of IS systems is Interactive Fiction (IF), in which players engage with a simulated environment through text-based \textit{player input} (or \textit{user input}). 
These games were remarkably sophisticated for their time, addressing world-modeling through complex rule sets grounded in formal linguistic structures~\cite{50_years_of_text_games}. 
Limited by their parsers, fixed action sets and a lack of common sense understanding, they often stifled player creativity and expression, constraining \textit{player agency}.
Large Language Models (LLMs) could address these challenges and revolutionize Human-Machine co-creation \cite{ferreira2026co}, yet they still struggle with narrative pace and logical consistency~\cite{skill_check,tian-etal-2024-large-language,teleki-etal-2025-survey}.
These limitations highlight a critical need for methods that ground LLM generation in more stable structures.

In this paper, we present a pilot study of world-state \textit{transformations}, a generalization of the classic IF actions that are suggested by LLMs and symbolically executed in a previously-defined structured world state (see Figure~\ref{fig:transformations_overview}).
This approach may provide a framework for Interactive Storytelling systems that aim to allow players to express themselves through their own creativity and exhibit some degree of improvisation.
We first conceptually discuss \textit{transformations}, followed by the methodological design of our pilot study and the analysis of the game logs. 
Finally, we present the key observations and findings from the evaluation.
The source code is available on GitHub\footnote{https://github.com/sgongora27/transformations}.

\section{Previous Work on the World-update Problem}

The main topic of this work is the \textbf{world-update} problem in IF systems, and text-based Interactive Storytelling in general.
Although multiple definitions exist~\cite{Riedl_Bulitko_2012,martin2016improvisational,hausknecht2020interactive,callison-burch-etal-2022-dungeons,skill_check}, the problem may be defined at a high level as follows:

\begin{center}
$action\ description\ (world\ state_n) \longrightarrow world\ state_{n+1}$
\end{center}

\noindent where $n \in \mathbb{N}$ represents a point in time of the represented fictional world. 
That is, $world\ state_{n+1}$ is a consequence of doing the described $action$ in the $world\ state_{n}$\footnote{Broadly speaking, it may happen that the described action has no impact in $world\ state_{n}$, in which case $world\ state_{n+1}=world\ state_{n}$.}.

Besides the parsers used in classic IF games like Zork---triggering a world-update process that was specifically pre-programmed for each available action~\cite{50_years_of_text_games}---many recent works (e.g., AI Dungeon\footnote{https://aidungeon.com/}) tried using LLMs for this purpose~\cite{Tuul_2023,dnd_emotions_2024}.
The aforementioned cases are representative for two very different strategies to tackle this problem, but there are many other ways to do it.
\citesubject{xie-etal-2025-making} represent the \textit{world state} through simple natural language sentences and use two fine-tuned LLMs to process the user input: one to detect invalid actions and another to generate outcomes for valid ones.
Neuro-symbolic approaches (i.e. those combining artificial neural networks, like LLMs, and symbolic approaches) have also been explored to tackle this problem.
\citesubject{callison-burch-etal-2022-dungeons} fine-tuned a language model that tracks the game state and generates a response based on a it.
One of their main conclusions is that additional machinery would be needed to track the game state beyond a Language Model.

\subsection{LLMs Predicting a Change in the World State}

Within a brief period, some works have proposed using LLMs to update structured world-state representations from free-text user input.

PAYADOR~\cite{gongora2024payador} is an approach to the world-update problem that tries to break with the tradition of using specific pre-programmed actions like in classic IF games, and focuses on modeling the general outcome of those actions instead.
This conceptual shift draws directly from modeling Tabletop Role-playing Games (TTRPG) and how Game Masters (GM) reason about the world state in response to player actions: rather than relying on an objective methodology for every possible action (something impossible given the open-ended nature of these games) they evaluate how a player's intent within its unique context will impact the fictional world~\cite{tychsen2005game}.

While PAYADOR was designed as a system-agnostic GM model, shortly thereafter, \citesubject{song2024you} introduced a specialized GM model for ``Jim Henson’s Labyrinth: The Adventure Game''.
In this model, an LLM maintains an ad-hoc structured representation through function calls.
In a contemporaneous effort, \citesubject{wang-etal-2024-language} also addressed LLM reasoning over world-state updates, but focusing on realistic environments.
While trying to check if LLMs can act as a reliable world model, they found sometimes this strategy can work better than predicting the whole world state each time.

\section{World-state Transformations}

The previously described works lead us to define what we call \textbf{world-state transformations}.
\textit{Transformations} are pre-programmed routines designed to symbolically update a world-state.
While retaining the spirit of classic IF, they aim to funnel a wide range of player actions, including the unpredictable ones born from creative expression. 
For instance, a player might use a standard medkit or spell\footnote{Classic IF actions a game designer typically considers.}, or they could improvise by creating or bartering for a remedy: an interaction too context-dependent for hard-coded logic to anticipate.
Such cases demand flexibility from the resulting IS system.
\textit{Transformations} may provide it, allowing LLMs to decide which of the available ones apply after any player input.
Moreover, a single input may trigger one or more \textit{transformations}, which collaborate together to shape the fictional world through synergic changes.

Classic approaches to the world-update problem triggered changes based on specific actions, limiting the player's options to a fixed set of outcomes predefined by the designers.
Instead, \textit{transformations} focus on general changes in the world state, regardless of the specific actions that trigger them.
As a middle ground between purely neural systems and classic approaches, this shift in focus enables pre-programmed updates to the world state while allowing players to express themselves through their own vocabulary and unique ideas.
As we will later discuss, \textit{transformations} seem to have enabled our evaluators to follow this direction.

Figure~\ref{fig:transformations_overview} illustrates how LLM-suggested \textit{transformations} may be used in a rule-based IS system. 
First, the symbolic IS system is built using an ad hoc representation that includes pre-programmed \textit{transformations}, each having additional pre-programmed consistency checks.
Then, every turn, an LLM receives the symbolic $world\ state$ (or a textual rendering of it), the free-text $player\ input$, and the list of available \textit{transformations}, along with instructions on how to suggest them. 
After the LLM determines which \textit{transformations} are triggered by the $player\ input$, the symbolic world model executes them subject to the corresponding consistency checks.

\section{Methodology}

As previously noted, both \citesubject{gongora2024payador} and \citesubject{song2024you} utilized world-state transformations within their respective IS systems. 
However, while the former did not include an empirical evaluation, the latter focused exclusively on human analysis of LLM-simulated sessions.
Consequently, our objective was to conduct a pilot study with humans using a live system (allowing for real-time interaction) to address three research questions regarding \textit{transformations}:

\begin{itemize}
    \itemsep -0.1em
    \item Are LLMs capable of correctly suggesting \textit{transformations} when grounded on a structured world state?
    \item Are \textit{transformations} suitable to tackle the world-update problem?
    \item Do \textit{transformations} encourage more creative player inputs?
\end{itemize}

Using the PAYADOR source code\footnote{{https://github.com/pln-fing-udelar/payador}} we designed two scenarios and tested them with eight players. 
We split the group evenly to cover two languages (English and Spanish) and two models (Gemini 1.5 Flash and Llama 3 70B). 

\section{System Description}

We built upon the original PAYADOR framework (see Table~\ref{tab:components} in the appendices for details) by integrating \texttt{Objectives}.
Each scenario can have one of the following objectives: the player is in a \texttt{Location} or has an \texttt{Item}; the player is in the same \texttt{Location} of another \texttt{Character}; or an \texttt{Item} is in a specific \texttt{Location}.
This was essential to conduct our experiments, as \texttt{Objectives} allow to establish clear completion criteria for our test scenarios, hence helping us determine if \textit{transformations} alone can act as a vehicle in human-machine communication.

\subsection{Considered World-state Transformations}

Inspired in the original PAYADOR source code, we implemented the following \textit{transformations}, each of them with robust consistency checks:

\begin{itemize}
    \item \textit{Moved Items} (MI-t): The LLM can suggest that an \texttt{Object} has to be moved from the player's inventory to the current \texttt{Location}, or vice-versa. It also includes the case of the player giving an NPC an \texttt{Object}, or vice-versa.
    \item \textit{Unblocked Locations (UL-t)}: The LLM can suggest that a blocked \texttt{Location} is now reachable.
    \item \textit{Player Movement (PM-t)}: The LLM can suggest that the player's \texttt{Character} has to be moved to a new \texttt{Location}.
\end{itemize}

Naturally, we could have included other \textit{transformations}.
However, to ensure a controlled experimental setting, we selected a minimal set of \textit{transformations} that align with traditional IF mechanics.
Given that we placed no constraints on how evaluators wrote their inputs, this decision narrows the range of possibilities during the experimental phase.

\section{Experimental Evaluation}

In this section we will present the two pre-designed scenarios and the results of each playthrough\footnote{All the game logs are available on GitHub: {https://github.com/sgongora27/transformations}}.
We manually examined every turn of each playthrough, tagging them according to two type of errors: \textbf{LLM} and \textbf{World Modeling} errors.

The first type is related to errors found when the \textbf{LLM} predicts the \textit{transformations} to be done: 
when the suggested \textit{transformations} do not correctly represent the outcomes of the \textit{user input}, or the \textit{transformations} are not correctly written by the LLM.
This error is further split into three categories, one for each \textit{transformation} (\textit{MI-t}, \textit{UL-t} and \textit{PM-t}).
In order to provide further insight into these cases, the world-update prompt and examples for text-based world rendering are included in the appendices.

The second type involves \textbf{world modeling} errors: cases where the LLM correctly proposes \textit{transformations}, but the final result fails to capture the player's intended actions or logical sequence due to system limitations.
This time we break this type of error into two categories.
The first, \textit{Planning}, refers to the execution order of \textit{transformations}. 
The system uses a default \textit{MI-t} $\rightarrow$ \textit{UL-t} $\rightarrow$ \textit{PM-t} sequence, so it fails when a player needs to do things in a different order.
The second one, \textit{memory}, refers to the lack of memory in PAYADOR, where the only information stored is the one in the structured representation, as the contents of the prompt are reset every turn. 
This means that the rest of details, such as previous dialogue utterances, are not saved. Some errors arise when the player refers to details that were not stored.

\begin{figure}[t!]
    \centering
    \includegraphics[width=\columnwidth]{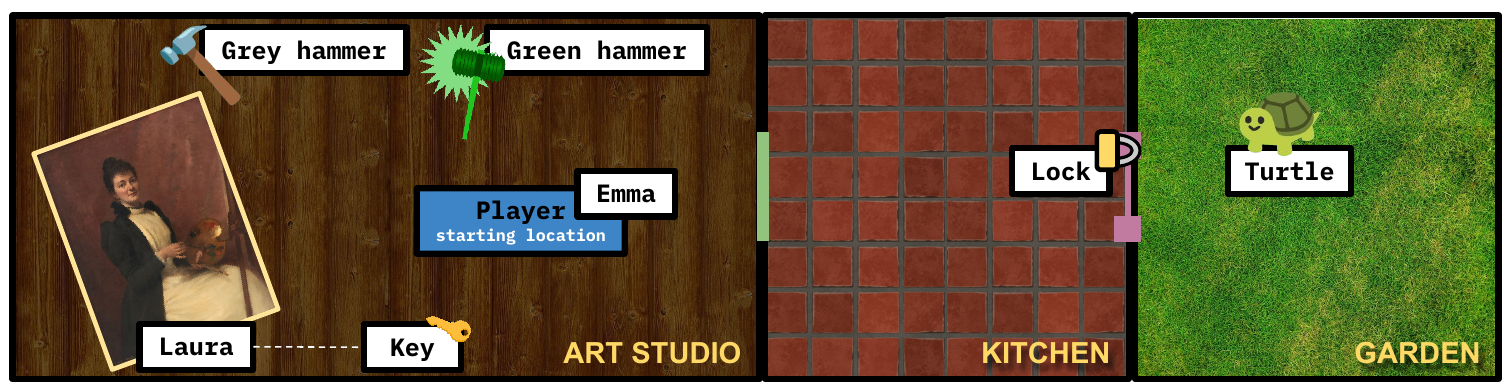}
    \caption{A graphical representation of Scenario A, including 2 \texttt{characters} (\textit{Laura} and \textit{Emma}), 5 \texttt{items} (\textit{Grey hammer}, \textit{Green hammer}, \textit{Key}, \textit{Lock} and \textit{Turtle}), and 3 \texttt{locations} (\textit{Art Studio}, \textit{Kitchen} and \textit{Garden}).}
    \label{fig:A_scenario_map_small}
\end{figure}

\subsection{Scenario A}

The first scenario consists of three \textit{locations}. 
The player acts as \textit{Emma}, a girl that has to find her \textit{Turtle} and take it to the \textit{Kitchen}.
Starting in her mother's \textit{Art Studio}, \textit{Emma} has to get the \textit{grey hammer} or the \textit{key} (from her mother, \textit{Laura}) to unlock the \textit{Garden} door, where the \textit{Turtle} is located.
This sequence requires frequent world-state updates using all three \textit{transformations}, aiming to intensely test them within a simple, realistic environment.
Figure~\ref{fig:A_scenario_map_small} illustrates the scenario; see the appendices for an enlarged version and the manual analysis results in Table~\ref{tab:errors_scenario_A}.

Overall, the eight testers managed to end the scenario. 
Some of them tried opening the lock with the \textit{key}, and some of them broke it using the \textit{hammer}.
In both languages, errors in the LLM's \textit{MI-t} suggestions were all traceable to an interesting issue: many players called the \textit{Turtle} \textit{Hojita} since the world state stores that ``Emma calls it \textit{Hojita}''
(e.g., ``I place Hojita in the kitchen, and walk away to the garden again, to enjoy the rest of this sunny day''-TesterD). 
While we expected the LLM to resolve this co-reference, it occasionally suggested an incorrect \textit{MI-t} using the name ``Hojita'', what was not recognized as a valid object by the world-modeling module, hence rejecting the \textit{transformation}.
Additionally, some \textit{PM-t} errors occurred because the player input was not explicit (e.g., ``As I walk through the door I look around trying to find my turtle, what do i see''-TesterC).
A single \textit{UL-t} error occurred when the LLM tried to unblock a location already shown as unblocked in its context.

Regarding world-modeling, many errors arose from the system's lack of \textit{planning}. 
Players often tried to move to the \textit{Kitchen} and drop the pet: an action requiring \textit{PM-t} before \textit{MI-t}. 
Our fixed execution order cannot accommodate this sequence, highlighting a limitation of the current default order.
Finally, there was a single error due to the memory of the system, when TesterE said ``agarro el otro martillo y me dirijo a la cocina'' (\textit{I grab the other hammer and head to the kitchen}), 
but the LLM could not disambiguate between the \textit{Grey hammer} and the \textit{Green hammer}.

\subsection{Scenario B}

The second scenario also consists of three \textit{locations}.
This time, the player acts as \textit{Venancio}, a \textit{gaucho}\footnote{According to the Cambridge Dictionary: \textit{a South American cowboy who is skilled at riding a horse}.} that has to find \textit{Artigas}\footnote{A historical figure of South America.}, who is locked in a \textit{Cell} protected by a \textit{Puzzle}.
In order to enter the \textit{Cell}, the player must first put out a \textit{firewall}\footnote{After running all the experiments we realized this component should be called \textit{Wall of Fire}. Apparently, the word \textit{firewall} generated no problems during the experiments in English.} and then solve a riddle.
To put out the \textit{firewall}, the player has to use \textit{Venancio}'s magic to summon a giant wave of water, what is indicated in his \textit{descriptions}.
In this case, the evaluation objective was twofold: first, to determine if LLM-predicted \textit{transformations} adhere to non-realistic magical physics; second, to assess the LLM's ability to describe critical scene elements, such as \textit{Venancio} summoning a water wave, ensuring players receive essential information to complete the scenario.
In the appendices, Figure~\ref{fig:B_scenario_map} illustrates this scenario, while Table~\ref{tab:errors_scenario_B} details its corresponding results.

As with Scenario A, all testers successfully completed the task.
Notably, world-modeling errors dropped to zero as a result of limited environmental interaction. 
This trend extended to the \textit{MI-t} category, where the sole error involved a consistency check rejecting TesterE's attempt to improvise an undeclared \texttt{Item}. 
However, this scenario yielded several notable \textit{UL-t} errors.
First, the particularity of having to solve a \texttt{Puzzle} to open the \textit{Cell} door introduced interesting cases.
Despite having the correct answer in-context, the LLM incorrectly validated TesterA and TesterC’s wrong riddle answers, suggesting a \textit{UL-t}, hence unblocking the \textit{Cell}. 
These cases illustrate how symbolic grounding mitigates LLM-dependent errors, but cannot entirely eliminate them.
Second, during TesterF's session, the LLM suggested an \textit{UL-t} for an already reachable place, mirroring TesterG’s experience in Scenario A.
Finally, while the wall of fire challenge sparked unconventional (and successful) ideas among the players, the LLM did not suggest a \textit{UL-t} when TesterB attempted to summon the wave by playing Venancio's guitar.

\section{Transformations and Co-Creativity}

Having discussed our observations from the qualitative analysis of errors, we will now explore the implications of \textit{transformations} for Interactive Storytelling systems.

We will start from the most salient fact: all players were able to complete the scenarios, both in English and Spanish.
Although our experiments are preliminary, the results seems to confirm that \textit{transformations} effectively bridge human-machine communication in LLM-enhanced IS systems.
 
Moreover, we observe that \textit{transformations} seem to foster player creativity.
For instance, TesterE unexpectedly used a window mentioned in the \textit{Kitchen}'s description to enter the \textit{Garden}, bypassing the intended \textit{Lock}.
More generally, players in both scenarios engaged by role-playing their characters and adding creative details to their inputs (e.g., speaking with \textit{Laura} in Scenario A or examining the magical setting in Scenario B).
This emergent capability may prove useful for systems requiring varying degrees of improvisation in human-machine co-creation.
Additionally, \textit{transformations} may be suitable for modeling TTRPGs, which scholars characterize as a form of organized, collaborative storytelling driven by rich player-GM interactions~\cite{mayra2017dialogue,TTRPG_creativity}

Finally, our results indicate no substantial disparity in performance between open-source models (Llama) and closed-source ones (Gemini). 
Performance also remained consistent across both English and Spanish language settings, which suggests current LLM are already capable enough of handling these scenarios.

While our experiments are preliminary, their results suggest that \textit{transformations} offer a promising path for consistent world updates in IS systems. 
Evidence indicates they broaden player expression by providing numerous options to trigger pre-programmed updates, making them ideal for improvisational systems like TTRPG models.
Although LLM errors may impact world coherence, the symbolic nature of \textit{transformations} facilitates the integration of logic to verify and refine outputs. 
Following the idea that improvisational storytelling is limited only by player imagination~\cite{martin2016improvisational}, our results suggest that \textit{transformations} represent a step forward in this direction.

\section{Conclusions}

In this paper, we evaluated world-state transformations as a method for tackling the \textit{world-update} problem in Improvisational Interactive Storytelling. 
By using an LLM to suggest pre-programmed world-state \textit{transformations}, the system moves away from manually defined actions toward \textbf{a model of general world changes}. 
This shift in focus encourages players to interact more creatively with the environment while maintaining the controllability inherent in symbolic world-state representations.

Our pilot study with eight participants across two scenarios confirmed the viability of this approach: all players successfully completed their tasks and engaged in co-creation while acting within the fictional world.
While the system inherits common LLM limitations, leading to errors like prematurely solved puzzles, the symbolic nature of the transformations allows for consistency checks that improve traceability and provide more transparent explainability than purely neural approaches.

Our experimental findings suggest several directions for further research. 
First, future work should explore creating world components from scratch, extending our recent preliminary efforts~\cite{vaucher2026ivie}, and combining transformations to create them on the fly, thereby better modeling the improvisation inherent in TTRPGs. 
As the number of transformations grows, incorporating a planner to dynamically order their execution will become essential. 
Finally, to minimize the impact of incorrect LLM outputs, we aim to optimize how consistency checks validate them.

We hope this work inspires other researchers to push the boundaries of human-machine co-creation in Interactive Storytelling, and the application of TTRPGs models to different domains.

\clearpage \pagebreak

\section{Author Contributions}
This work is part of S.G.’s Master’s thesis~\cite{gongora2025approaches}, for which L.C. and G.M. served as co-advisors. 
P.G. helped frame the conceptual aspects of this work.

\section{Acknowledgments}

This paper was partially funded by \textit{Agencia Nacional de Investigación e Innovación} (ANII, Uruguay), Grant No. $POS\_NAC\_2022\_1\_173659$; \textit{Comisión Académica de Posgrado} (CAP, Uruguay), Grant No. $BDDX\_2026\_1\#48948494$; and by the project DARK NITE: \textit{Dialogue Agents Relying on Knowledge-Neural hybrids for Interactive Training Environments}, Grant No. $PID2023-146308OB-I00$ (Spanish Ministry of Science
and Innovation).

\bibliographystyle{iccc}
\bibliography{iccc}

\clearpage \pagebreak 

\appendix
\section{Appendices}

\subsection{World-state components}

Table~\ref{tab:components} shows the four classes present in our system, strongly based on the PAYADOR source code (\texttt{Locations}, \texttt{Characters}, \texttt{Objects} and \texttt{Puzzles}).
All these components are tracked by a \texttt{World} class, representing the \textit{world state}.

\begin{table}[h!]
\footnotesize
\centering
\scalebox{0.85}{
    \begin{tabular}{|p{0.19\textwidth }|p{0.31\textwidth}|}
    \hline
    \textbf{Item}   & \textbf{Location}\\\hline
    \begin{tabular}[t]{@{}l@{}}
    \textit{Name}: \textbf{String}\\
    \textit{Descriptions}: List [\textbf{String}]\\
    \textit{Gettable}: \textbf{Boolean}
    \end{tabular} 
    & 
    \begin{tabular}[t]{@{}l@{}}
    \textit{Name}: \textbf{String}\\
    \textit{Descriptions}: List [\textbf{String}]\\
    \textit{Items}: List [\textbf{Item}]\\
    \textit{Connecting locations}: List [\textbf{Location}]\\
    \textit{Blocked locations}: List [\textbf{Location}, \textit{obstacle}]
    \end{tabular} 
    \\\hline \hline
    \textbf{Character} & \textbf{Puzzle}\\\hline
    \begin{tabular}[t]{@{}l@{}}
    \textit{Name}: \textbf{String}\\
    \textit{Descriptions}: List [\textbf{String}]\\
    \textit{Location}: \textbf{Location}\\
    \textit{Inventory}: List [\textbf{Item}]
    \end{tabular} 
    &
    \begin{tabular}[t]{@{}l@{}}
    \textit{Name}: \textbf{String}\\
    \textit{Descriptions}: List [\textbf{String}]\\
    \textit{Problem}: \textbf{String}\\
    \textit{Answer}: \textbf{String}
    \end{tabular} 
    \\\hline
    \end{tabular}
}
\caption{\label{tab:components} The attributes of each \texttt{World} component in our representation. The \textit{obstacle} of a blocked location can be any other component, but it is particularly thought to be an \texttt{Item} or a \texttt{Puzzle}.}
\end{table}

\subsection{Examples of rendered world-states}

\begin{promptbox}{Starting scene --- Scenario A (English)}
\begin{PromptVerbatim}
The player is in <Art studio>
From <Art studio> the player can access: <Kitchen>
From <Art studio> there are blocked passages to: None
The player has the following objects in the inventory: None
The player can see the following objects: <A grey hammer>, <A green hammer>
The player can see the following characters: <Laura>

Here is a description of each component.
<Art studio>: This is the player's location. This is the art studio that Emma's mom has in the house.
Characters:
- <Player>: The player is acting as <Emma>. A teenager of average height. She is looking for her pet 'Hojita'.
- <Laura>: A woman in her 40s. She is Emma's mom. She is an artist, and loves oil painting. This character has the following items: <Key>
Objects:
- <A grey hammer>: A big grey hammer that can be used to break things. It is so heavy...
- <A green hammer>: A small green hammer. It is just a toy and you cannot break anything with it
- <Key>: A key to open a lock. It is golden. There is a strange coat of arms engraved on it
\end{PromptVerbatim}
\end{promptbox}

\begin{promptbox}{Starting scene --- Scenario B (English)}
\begin{PromptVerbatim}
The player is in <Clearing in the woods>
From <Clearing in the woods> the player can access: None
From <Clearing in the woods> there are blocked passages to: <Silent zone> blocked by <Firewall>
The player has the following objects in the inventory: <Guitar>
The player can see the following objects: <Writings>, <Pond>
The player can see the following characters: None

Here is a description of each component.
<Clearing in the woods>: This is the player's location. A clearing in a eucalyptus forest near the Uruguay River. You can hear the sound of the animals that live in the trees of this forest..
Characters:
- <Player>: The player is acting as <Venancio>. A Uruguayan gaucho in his 40s. He belongs to the Artigas army. He has the magical power to summon a giant wave of water with which he can put out fires or moisten the ground..
Objects:
- <Writings>: There is something written on the wall.. It says 'You have to trust in the powers that have been given to you.'
- <Pond>: A pond full of crystal clear water. The water is so clear that it works like a mirror
- <Guitar>: A classic guitar with 6 strings. It sounds great
- <Firewall>: The flames are very hot. It's 3 metres high. It is impossible to cross them, neither walking, nor running, nor jumping.
\end{PromptVerbatim}
\end{promptbox}

\subsection{World-update prompt}

\begin{promptbox}{World-update prompt (English)}
\begin{PromptVerbatim}
system_msg = f"""You are a storyteller. You are managing a fictional world, and the player can interact with it. Following a specific format, that I will specify below, your task is to find the changes in the world after the actions in the player input. Specifically, you will have to find what objects were moved, which previously blocked passages are now unblocked, and if the player moved to a new place.
       
Here are some clarifications:
(A) Pay attention  to the description of the components and their capabilities.
(B) If a passage is blocked, then the player must unblock it before being able to reach the place. Even if the player tells you that he is going to access the locked location, you have to be sure that he is complying with what you asked to allow him to unlock the access, for example by using a key or solving a puzzle.
(C) Do not assume that the player input always makes sense; maybe those actions try to do something that the world does not allow.
(D) Follow always the following format with the three categories, using "None" in each case if there are no changes and repeat the category for each case:
- Moved object: <object> now is in <new_location>
- Blocked passages now available: <now_reachable_location>
- Your location changed: <new_location>
(E) Finally, you can narrate the changes you've detected in the world state (without moving the story forward and without making up details not included in the world state!) using the format: #your final message#
(F) In the narration section that you add at the end, between # symbols, you can also answer questions that the player asks in their input, about the objects or characters they can see, or the place they are in.

Here I give you some examples (in parentheses, a clarification about what the player might have tried to do) for the asked format, as described in items (D) and (E):

Example 1 (The player took the axe and put it in the inventory)
- Moved object: <axe> now is in <Inventory>
- Blocked passages now available: None
- Your location changed: None
#You put the axe in your bag#

Example 2 (The player unblocks the passage to the basement)
- Moved object: None
- Blocked passages now available: <Basement>
- Your location changed: None
# The basement is now reachable #

Example 3 (The player now is in the garden)
- Moved object: None
- Blocked passages now available: None
- Your location changed: <Garden>
# You enter the garden #

Example 4 (The player puts objects in the bag and leaves the axe on the floor)
- Moved object: <banana> now is in <Inventory>,  <bottle> now is in <Inventory>,  <axe> now is in <Main Hall>
- Blocked passages now available: None
- Your location changed: None
# You put the banana and the bottle in your bag. The axe lies on the floor of the Main hall #

Example 5 (The player puts objects in the bag and leaves the axe on the floor and unblocks the passage to the Small room)
- Moved object: <banana> now is in <Inventory>,  <bottle> now is in <Inventory>,  <axe> now is in <Main Hall>
- Blocked passages now available: <Small room>
- Your location changed: None
# You put the banana and the bottle in your bag. The axe lies on the floor of the Main hall. Now you can reach the Small room. #

Example 6 (The player puts objects in the bag and leaves the axe on the floor, unblocks the passage and goes to the Small room)
- Moved object: <banana> now is in <Inventory>,  <bottle> now is in <Inventory>,  <axe> now is in <Main Hall>
- Blocked passages now available: <Small room>
- Your location changed:  <Small room>
# You put the banana and the bottle in your bag. The axe lies on the floor of the Main hall. The Small room is now unblocked, and you moved there. #

Example 7 (The player puts the pencil in the bag and gives the book to John)
- Moved object: <book> now is in <John>,  <pencil> now is in <Inventory>
- Blocked passages now available: None
- Your location changed:  None
# John now has the book. You put the pencil in your bag #

Example 8 (The player gives the computer to Susan)
- Moved object: <computer> now is in <Susan>
- Blocked passages now available: None
- Your location changed:  None
# Susan put the computer in her bag #

Example 9 (The player does something that has not the expected outcome)
- Moved object: None
- Blocked passages now available: None
- Your location changed:  None
# Nothing happened... #

Example 10 (The player asks a question)
- Moved object: None
- Blocked passages now available: None
- Your location changed:  None
# Answer to the player's question #"""

user_msg = f"""Give the changes in the world following the specified format, after this player input "\{input\}" on this world state: \{world_state\}"""
\end{PromptVerbatim}
\end{promptbox}

\subsection{Scenario results}

In this appendix we include tables with the results of the manual analysis of the playthroughs.
These tables (Table~\ref{tab:errors_scenario_A} and Table~\ref{tab:errors_scenario_B}) show LLM-related errors alongside errors stemming from symbolic world-modeling limitations.

\begin{table}[h!]
\small
\centering
\scalebox{0.87}{
\begin{tabular}{|l|c|c|c|c|c|}
\hline
\textbf{Scenario A}  & \multicolumn{3}{c|}{\textbf{LLM Errors}} & \multicolumn{2}{c|}{\textbf{World Modeling Errors}} \\ \cline{2-6} 
 & \textbf{MI-t} & \textbf{PM-t} & \textbf{UL-t} & \textbf{Planning} & \textbf{Memory} \\ \hline
\textbf{TesterA} (Gemini) & 0 & 0 & 0 &  1 & 0  \\ \hline
\textbf{TesterB} (Gemini) & 0 & 0 & 0 &  1 & 0  \\ \hline
\textbf{TesterC} (Llama)  & 3 & 1 & 0 &  0 & 0  \\ \hline
\textbf{TesterD} (Llama)  & 4 & 1 & 0 &  2 & 0  \\ \hline
\textbf{Total} (English)  & 7 & 2 & 0 &  4 & 0   \\ \hline \noalign{\smallskip} \hline
\textbf{TesterE} (Gemini) & 3 & 0 & 0 &  2 & 1 \\ \hline
\textbf{TesterF} (Gemini) & 1 & 0 & 0 &  0 & 0 \\ \hline
\textbf{TesterG} (Llama)  & 4 & 0 & 1 & 1 & 0 \\ \hline
\textbf{TesterH} (Llama)  & 0 & 0 & 0 & 0 & 0 \\ \hline
\textbf{Total} (Spanish)  & 8 & 0 & 1 & 3 & 1 \\ \hline
\end{tabular}
}
\caption{Number of errors by category for each of the eight testers during the playthroughs of Scenario A. Testers A-D played in English, while Testers E-H played in Spanish. The LLM used is indicated for each case, next to the tester's id.}
\label{tab:errors_scenario_A}
\end{table}

\begin{table}[h!]
\small
\centering
\scalebox{0.87}{
\begin{tabular}{|l|c|c|c|c|c|}
\hline
\textbf{Scenario B}  & \multicolumn{3}{c|}{\textbf{LLM Errors}} & \multicolumn{2}{c|}{\textbf{World Modeling Errors}} \\ \cline{2-6} 
 & \textbf{MI-t} & \textbf{PM-t} & \textbf{UL-t} & \textbf{Planning} & \textbf{Memory} \\ \hline
\textbf{TesterA} (Gemini) & 0 & 0 & 1 & 0 & 0  \\ \hline
\textbf{TesterB} (Gemini) & 0 & 0 & 1 & 0 & 0  \\ \hline
\textbf{TesterC} (Llama)  & 0 & 0 & 1 & 0 & 0  \\ \hline
\textbf{TesterD} (Llama)  & 0 & 0 & 0 & 0 & 0  \\ \hline
\textbf{Total} (English)  & 0 & 0 & 3 & 0 & 0 \\ \hline \noalign{\smallskip} \hline
\textbf{TesterE} (Gemini) & 1 & 1 & 2 & 0 & 0 \\ \hline
\textbf{TesterF} (Gemini) & 0 & 2 & 1 & 0 & 0 \\ \hline
\textbf{TesterG} (Llama)  & 0 & 1 & 0 & 0 & 0 \\ \hline
\textbf{TesterH} (Llama)  & 0 & 1 & 0 & 0 & 0 \\ \hline
\textbf{Total} (Spanish) &  1 & 5 & 3 & 0 & 0 \\ \hline
\end{tabular}
}
\caption{Number of errors by category for each of the eight testers during the playthroughs of Scenario B. Testers A-D played in English, while Testers E-H played in Spanish. The LLM used is indicated for each case, next to the tester's id.}
\label{tab:errors_scenario_B}
\end{table}

\subsection{Scenario illustrations}

\begin{figure*}[h!]
    \centering
    \includegraphics[width=\linewidth]{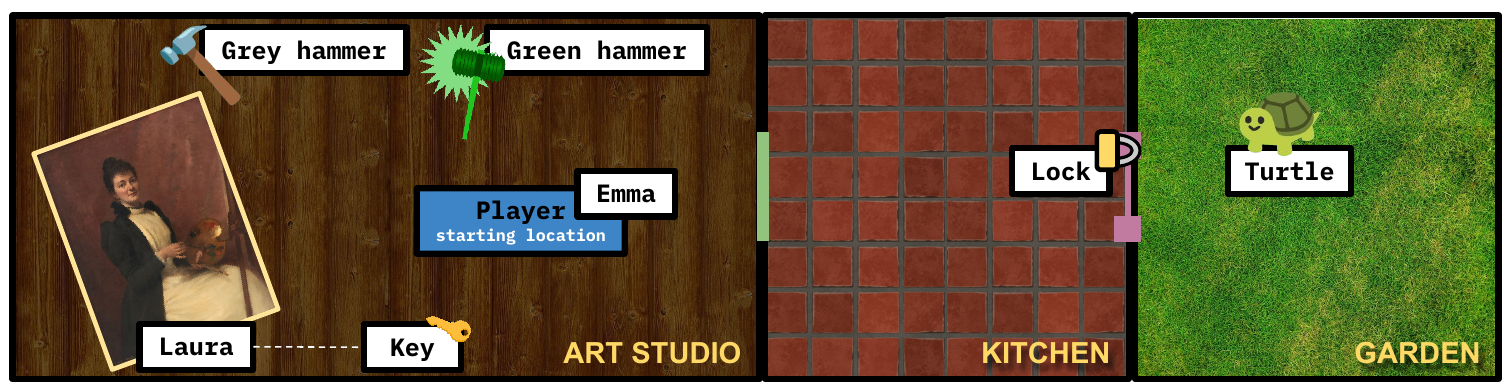}
    \caption{A graphical representation of Scenario A. Ten \texttt{world} components are represented: 2 \texttt{characters} (\textit{Laura}, an NPC, and \textit{Emma}, the player character), 5 \texttt{items} (\textit{Grey hammer}, \textit{Green hammer}, \textit{Key}, \textit{Lock} and \textit{Turtle}, and 3 \texttt{locations} (\textit{Art Studio}, \textit{Kitchen} and \textit{Garden}).}
    \label{fig:A_scenario_map}
\end{figure*}

\begin{figure*}[h!]
    \centering
    \includegraphics[width=\linewidth]{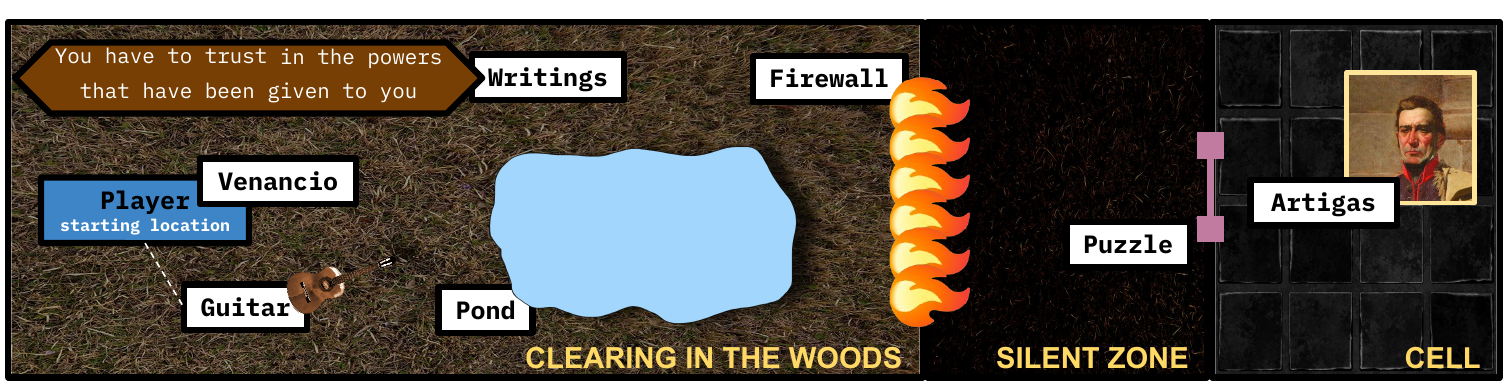}
    \caption{A graphical representation of Scenario B. Ten \texttt{world} components are represented: 2 \texttt{characters} (\textit{Artigas}, an NPC, and \textit{Venancio}, the player), 4 \texttt{items} (\textit{Guitar}, \textit{Pond}, \textit{Writings}, and \textit{Firewall}), 3 \texttt{locations} (\textit{Clearing in the woods}, \textit{Silent zone} and \textit{Cell}), and a \texttt{puzzle} (\textit{Puzzle})}
    \label{fig:B_scenario_map}
\end{figure*}

In this appendix we include illustrations to help visualizing the pre-made scenarios we designed to test \textit{Transformations}.
These figures (Figure\ref{fig:A_scenario_map} and Figure\ref{fig:B_scenario_map}) are for illustrative purposes only; players interacted with the environment via a text-only interface.

\subsection{A real example using Gemini}

\begin{figure*}[t!]
    \centering
    \includegraphics[width=\textwidth]{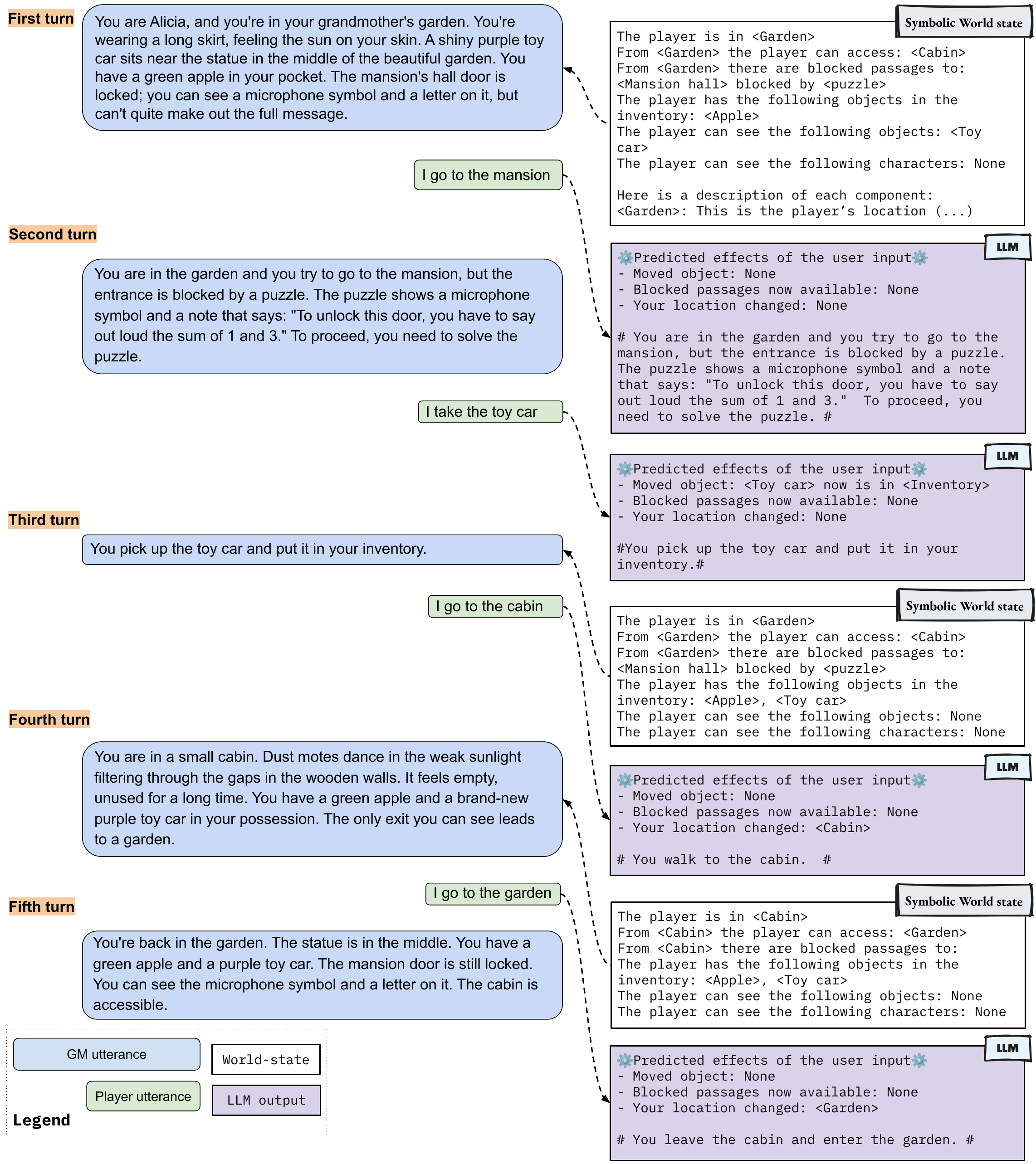}
    \caption{A five-turn gameplay example using Gemini 1.5 Flash. The left panel displays the dialogue between the automated Game Master and the human player, while the right panel shows the symbolic world states (for representative turns) and the LLM-suggested \textit{transformations} (for all turns). A legend is provided in the bottom left-hand corner.}
    \label{fig:payador_example}
\end{figure*}

Figure~\ref{fig:payador_example} shows a real example using Gemini 1.5 Flash, with step-by-step annotations of human utterances and LLM outputs (both GM utterances and suggested \textit{transformations}).

\clearpage \pagebreak

\end{document}